\documentclass[letterpaper,10pt,conference]{ieeeconf}
\IEEEoverridecommandlockouts                              

\usepackage{cite}
\usepackage{amsmath,amssymb,amsfonts}
\usepackage{subcaption}
\usepackage{textcomp}
\usepackage{xcolor}

\usepackage{graphicx}
\graphicspath{{figures}}
\DeclareGraphicsExtensions{.pdf,.eps,.jpeg,.png}

\usepackage{booktabs}       
\usepackage{multirow}
\usepackage{makecell}
\usepackage{subcaption}

\usepackage[breaklinks=true]{hyperref}

\usepackage{soul}
\usepackage{bm}

\newcommand{\mysubsection}[1]{{\bf#1.}~}

\usepackage{changes}
\makeatletter
\AddToHook{cmd/added/before}{\def\Changes@AuthorColor{red}}
\AddToHook{cmd/deleted/before}{\def\Changes@AuthorColor{blue}}
\makeatother

\title{\Large \bf Improved 3D Gaussian Splatting of Unknown Spacecraft Structure Using Space Environment Illumination Knowledge}

\author{Tae Ha Park\textsuperscript{1} ~~~ Simone D'Amico\textsuperscript{2}%
\thanks{T.~H.~Park is supported by the Technological Innovation R\&D Program (RS-2024-00513881) funded by the Ministry of SMEs and Startups, Republic of Korea.}%
\thanks{\textsuperscript{1}Nara Space Technology Inc., Seoul, Republic of Korea. Email: {\tt\small thpark@naraspace.com}}%
\thanks{\textsuperscript{2}Dept.~of Aeronautics \& Astronautics, Stanford University, Stanford, CA 94305, USA. Email: {\tt\small damicos@stanford.edu}}%
}

\begin{document}

\maketitle
\bstctlcite{IEEEexample:BSTcontrol}

\begin{abstract}
This work presents a novel pipeline to recover the 3D structure of an unknown target spacecraft from a sequence of images captured during Rendezvous and Proximity Operations (RPO) in space. The target's geometry and appearance are represented as a 3D Gaussian Splatting (3DGS) model. However, learning 3DGS requires static scenes, an assumption in contrast to dynamic lighting conditions encountered in spaceborne imagery. The trained 3DGS model can also be used for camera pose estimation through photometric optimization. Therefore, in addition to recovering a geometrically accurate 3DGS model, the photometric accuracy of the rendered images is imperative to downstream pose estimation tasks during the RPO process. This work proposes to incorporate the prior knowledge of the Sun's position, estimated and maintained by the servicer spacecraft, into the training pipeline for improved photometric quality of 3DGS rasterization. Experimental studies demonstrate the effectiveness of the proposed solution, as 3DGS models trained on a sequence of images learn to adapt to rapidly changing illumination conditions in space and reflect global shadowing and self-occlusion.
\end{abstract}


\section{Introduction} \label{sec:intro}

Autonomous Guidance, Navigation and Control (GN\&C) with respect to a non-cooperative spaceborne target is a core technology indispensable to various Rendezvous and Proximity Operation (RPO) missions, such as on-orbit servicing \cite{reed_aiaaspace_2016_restorel, pyrak_2022_amos_mev} and active debris removal \cite{aglietti_2019_aer_removedebris}. In particular, monocular vision-based approaches have received widespread attention from the community due to their reliance on a low Size, Weight, Power and Cost (SWaP-C) optical sensor suitable for on-board satellite avionics. In this case, the servicer spacecraft must be able to estimate in real-time the position and orientation (i.e., \emph{pose}) of the target object with respect to the servicer given a sequence of images captured during the RPO process.

A predominant body of existing literature on the topic of vision-based spacecraft pose estimation assumes known targets whose 3D structure and mass properties are available a priori to the operators \cite{damico_2014_ijsse_pose, sharma_2018_jsr_robust, capuano_2019_scitech_robust, grompone_2015_phd, petit_2011_iros_pose}. In this setting, many works adopted Machine Learning (ML) and Neural Networks (NN) to perform pose estimation from a single image \cite{kisantal_2020_taes_spec, park_2023_acta_spec2021, park_2019_aas_krn, park_2024_asr_spnv2, park_2024_spnv3, sharma_2018_aero_pose, sharma_2020_taes_spn, chen_2019_iccvw_pose, proenca_2020_icra_urso, pasqualetto_2021_acta_cnn, hu_2021_cvpr_swisscube, garcia_2021_cvprw_lspnet} or continuous tracking with a navigation filter in the loop given a sequence of images during RPO \cite{park_2023_jgcd_spnukf, park_2024_icra_ost, pasqualetto_2023_acta_ukf, kaki_2023_jais_pose}. Due to the inaccessibility of space, these methods primarily leverage computer-rendered synthetic images for training, available via various open-source datasets such as SPEED \cite{sharma_2020_taes_spn}, SPEED+ \cite{park_2022_aero_speedplus}, URSO \cite{proenca_2020_icra_urso} and SPARK \cite{musallam_2021_icipc_spark}. Moreover, in order to prevent overfitting to the synthetic imagery and bridge the performance gap when tested on previously unknown spaceborne imagery, the robustness and generalizability of the CNN models are validated on-ground using hardware-in-the-loop testbeds capable of emulating space-like illumination conditions \cite{park_2021_aas_tron, park_2022_aero_speedplus, pasqualetto_2022_acta_ogrl, olivaresmendez_2023_jsse_zeroglab}.

\begin{figure}[!t]
    \centering
    \includegraphics[width=0.48\textwidth]{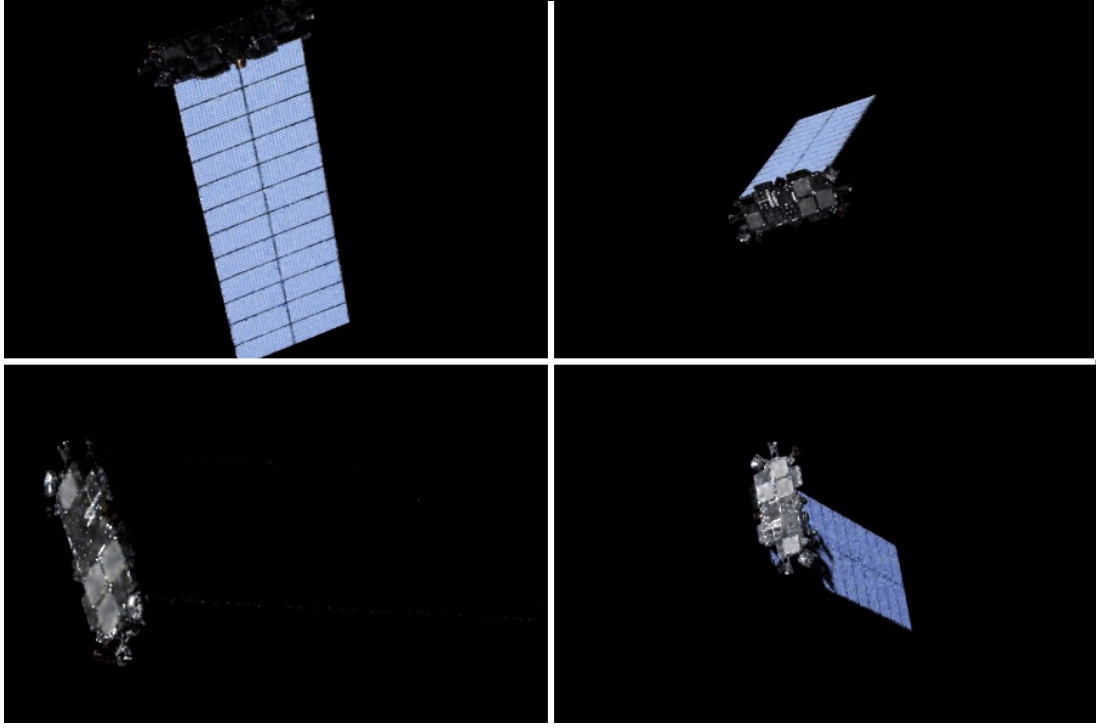}
    \caption{The rendered images of a previously unknown satellite generated by a 3DGS model trained on a sequence of images captured during a simulated RPO scenario. Compared with vanilla 3DGS, the proposed training pipeline explicitly incorporates knowledge of the Sun vector maintained by the onboard AOCS module through shadow splatting, significantly improving the photometric accuracy of the rendered images.}
    \label{fig:headline}
\end{figure}

In order to support a full range of future RPO missions including debris removal, the assumption of known targets must be relaxed. This means that the servicer spacecraft must be capable of not only predicting the target's motion but also reasoning about the target's 3D geometry and inertial properties simultaneously. If these are all done in an online sequence, the problem is akin to the well-known Simultaneous Localization and Mapping (SLAM) \cite{murartal_2015_robotics_orbslam} in modern robotics. 

Current state-of-the-art approaches to visual SLAM adopt radiance field models such as Neural Radiance Fields (NeRF) \cite{mildenhall_2020_eccv_nerf} and 3D Gaussian Splatting (3DGS) \cite{kerbl_2023_acm_3dgs} as a unified representation of an unknown scene. Radiance field models enable high-fidelity novel view synthesis by learning to map the camera's position and viewing direction to view-dependent color and density of the scene via differentiable volumetric rendering. While the original designs of NeRF and 3DGS required known pose labels acquired from the initial Structure-from-Motion (SfM) stage \cite{schonberger_2016_cvpr_colmap, schonberger_2016_eccv_colmap}, recent works have shown that these models can be incorporated into the SLAM framework and learned without prior knowledge of the ground-truth camera poses \cite{rosinol_2023_iros_nerfslam, barad_2024_isparo_3dgs, fu_2024_cvpr_colmapfree3dgs, matsuki_2024_cvpr_gsslam}. Specifically, the poses can be recovered by setting them as independent variables in the optimization of the photometric loss for rendered images while fixing the 3DGS parameters. Naturally, the photometric accuracy of differentiable rendering is crucial to the recovery of a scene or object in a visual SLAM scenario.

However, original NeRF and 3DGS pipelines assume controlled static scenes with consistent illumination conditions, as the scene colors are represented as spherical harmonic coefficients dependent only on the camera's viewing directions. Such a setting is not suitable for a space environment, where directional light of the Sun can cause significant and rapidly varying occlusion on the target RSO as its orbital position and attitude change over time. A conventional approach to learning NeRF or 3DGS in this so-called ``in-the-wild'' setting is to assign a learnable appearance embedding vector to each training image, effectively capturing the transient components of the scene (e.g., weather, time, etc.) into the low-dimensional latent space \cite{martinbrualla_2021_cvpr_nerfw, kulhanek_2024_nips_wildgaussians}. However, there are two issues with the direct application of in-the-wild methods to RPO scenarios. First, the correct appearance embedding for the new images must be solved at test-time via multiple rounds of computationally expensive differentiable rasterization. Second, the images acquired during RPO are likely to be quickly discarded due to limited on-board computational power, which prevents the proper training of per-image appearance embeddings and the downstream 3DGS pipeline.

In light of these challenges, this paper proposes a novel 3DGS pipeline that can recover the geometry and appearance of an unknown target RSO despite rapidly changing illumination conditions in space. Specifically, as opposed to conventional in-the-wild methods which require test-time optimization of per-image appearance embeddings, this work proposes to leverage the prior information that every self-navigating satellite possesses---the Sun vector. Considering that the Sun is the predominant source of illumination in space and that the spacecraft maintains an accurate estimate of its direction for its own Attitude and Orbit Control System (AOCS), incorporating its knowledge into the in-the-wild 3DGS pipeline results in an improved performance without having to optimize the embedding. Specifically, inspired by recent literature on relightable Gaussian Splatting \cite{bi_2024_siggraph_gs3}, the visibility of each Gaussian due to direct lighting is explicitly obtained by splatting the Gaussians towards the known lighting direction. The resulting visibility/shadow information is provided as an additional input to the pipeline, accelerating the photometrically accurate recovery of the observed images. As previewed in Fig.~\ref{fig:headline}, experiments on high-fidelity synthetic images rendered during representative RPO scenarios suggest that the proposed pipeline can recover the geometrically and photometrically accurate 3DGS models that capture global illumination effects and self-occlusion.

\section{Related Work}

\mysubsection{Spaceborne Visual SLAM}
Early studies for vision-based navigation about an unknown target RSO leveraged existing feature detection algorithms and SLAM frameworks. For example, Tweddle \cite{tweddle_2013_thesis} utilized SURF features \cite{bay_2008_surf} and iSAM \cite{kaess_2008_robotics_isam} to track the motion of a SPHERES satellite within the controlled environment inside the International Space Station (ISS). Dor \& Tsiotras \cite{dor_2018_sfm_orbslam} applied ORB-SLAM \cite{murartal_2015_robotics_orbslam} to real images of the Hubble Space Telescope (HST) captured during the NASA STS-125 Servicing Mission 4, but they had to recover its scale using the ground-truth model of HST.

More recent works investigate the applicability of ML to this problem. While not exactly a SLAM method, Park \& D'Amico \cite{park_2024_scitech_spe3r} proposed an \emph{offline} training of a Convolutional Neural Network (CNN) to reconstruct an unknown spacecraft structure as a set of superquadric primitives \cite{barr_1981_superquadric} from single images. The model is trained on the SPE3R dataset comprising high-fidelity synthetic images of 64 different spacecraft models. Naturally, the resulting model overfits to SPE3R and lacks zero-shot capability when subject to images of a completely new spacecraft. Moreover, the predicted orientation often exhibits large errors due to the symmetric nature of man-made spacecraft. Follow-up works \cite{park_2024_iwscff_sq, bates_2025_gnc_ambiguity} attempted to address the aforementioned shortcomings with architectural innovation and accounting of pose ambiguities. Specifically, Bates \& D'Amico \cite{bates_2025_gnc_ambiguity} proposed an ambiguity-free solution by redefining the target's body axes to be parallel to the camera axes for each training instance, effectively coupling the 3D shape and orientation estimations. This approach resulted in an improved 3D reconstruction quality on unseen spacecraft models without explicit consideration of their geometric symmetry.

On the other hand, a number of works attempted to leverage radiance field models for vision-based spaceborne navigation about an unknown RSO. Caruso et al.~\cite{caruso_2023_sfm_nerf} and Mergy et al.~\cite{mergy2021_cvprw_nerf} implemented various NeRF models \cite{muller_2022_acm_instantngp, pumarola_2021_cvpr_dnerf} for 3D reconstruction of an unknown spacecraft from a set of synthetic and hardware images. However, they still rely on COLMAP \cite{schonberger_2016_cvpr_colmap, schonberger_2016_eccv_colmap} to recover the initial pose labels. Recently, Barad et al.~\cite{barad_2024_isparo_3dgs} proposed an object-centric SLAM pipeline based on 3DGS representation of the target, whereby a small window of keyframes is maintained to train both 3DGS parameters and camera poses in an interchanging fashion. Tested on sequences of synthetic images rendered during the simulations of a spiral trajectory about 10 different spacecraft models, the experiment showed promising results but failed to account for photometric variation due to dynamic illumination conditions. 

\mysubsection{Radiance Field Models in Dynamic Scenes}
As noted earlier, vanilla radiance field models assume static scenes and thus have difficulty by design accounting for different dynamic elements such as lighting conditions, weather, and transient objects such as moving pedestrians. NeRF-W \cite{martinbrualla_2021_cvpr_nerfw} is a pioneering approach that abstracts away per-image photometric and environmental variations and transient components into separate learnable low-dimensional latent spaces. The core principle of disentangling static and dynamic components is adopted by many NeRF- and 3DGS-based works that followed \cite{tancik_2022_cvpr_blocknerf, yang_2023_iccv_crossraynerf, chen_2022_cvpr_ha-nerf, kulhanek_2024_nips_wildgaussians, zhang_2024_eccv_gs-w, dahmani_2024_eccv_swag}. However, as explained earlier, conventional in-the-wild methods are not compatible with RPO scenarios in space due to the need for finding the optimal appearance latent vector for each evaluation image.

\section{Methodology} \label{sec:method}

The proposed pipeline aims to perform \emph{online} training of a 3DGS model that can render geometrically and photometrically accurate images of the target RSO given known directions to the Sun as estimated by the servicer spacecraft. As noted in Sec.~\ref{sec:intro}, the photometric accuracy of 3DGS models is extremely important since the unknown camera poses can be recovered by optimization using the photometric loss function on the rendered images obtained via differentiable rendering. Therefore, in order to limit the scope of this paper to the photometric accuracy of 3DGS rasterization in spaceborne RPO scenarios, a number of favorable assumptions are made throughout this paper, including known pose labels and availability of binary masks to remove the background. Moreover, instead of relying on COLMAP \cite{schonberger_2016_cvpr_colmap, schonberger_2016_eccv_colmap}, the initial Gaussian positions are uniformly sampled from the surfaces of the ground-truth RSO 3D models. The ways to relieve these ideal assumptions are detailed later in Sec.~\ref{sec:discussions}.

\mysubsection{Preliminary: 3D Gaussian Splatting}
3D Gaussian Splatting (3DGS) \cite{kerbl_2023_acm_3dgs} represents the target or scene as a sparse set of anisotropic 3D Gaussians, where the $i^\text{th}$ Gaussian $\mathcal{G}_i$ is defined by a 3D covariance matrix $\bm{\Sigma}_i \in \mathbb{S}_+^3$ centered at mean $\bm{\mu}_i \in \mathbb{R}^3$:
\begin{align}
    \mathcal{G}_i(\bm{x}) = \exp \bigg[ -\frac{1}{2} (\bm{x} - \bm{\mu}_i)^\top \bm{\Sigma}_i^{-1} (\bm{x} - \bm{\mu}_i) \bigg]
\end{align}
Note that its mean and covariance respectively represent the position and 3D ellipsoidal shape of $\mathcal{G}_i$ in the RSO's body frame. They are additionally equipped with optical properties including view-dependent RGB color $\bm{c}_i$ represented as Spherical Harmonic (SH) coefficients and opacity $o_i \in [0, 1]$.

The aforementioned geometric and optical parameters of 3D Gaussians are learned using a differentiable rasterizer and gradient-based optimization. During forward propagation, the 3D Gaussians are ``splatted'' onto 2D image space via projective transformation. In order to make the process differentiable, the splatted 2D covariance $\bm{\Sigma}^\text{S} \in \mathbb{S}_+^2$ is obtained by ignoring the third row and column after an affine approximation of the projective transformation \cite{zwicker_2001_vis_ewasplatting} given as
\begin{align}
    \bm{\Sigma}^\text{S} = \big( \bm{J} \bm{W} \bm{\Sigma} \bm{W}^\top \bm{J}^\top \big)_{1:2,1:2}
\end{align}
where $\bm{J}$ is the Jacobian of the projective transformation and $\bm{W}$ is the viewing matrix. This approximation allows for numerical back-propagation of the photometric loss functions to the Gaussian parameters. Note that directly optimizing the covariance matrix $\bm{\Sigma}$ does not necessarily preserve its positive semi-definiteness. Therefore, $\bm{\Sigma}$ is further decomposed as $\bm{\Sigma} = \bm{R} \bm{S} \bm{S}^\top \bm{R}^\top$ with an identity scale matrix $\bm{S}$ and a rotation matrix $\bm{R}$, which is in turn represented as a unit-norm quaternion vector.

Once the Gaussians are splatted onto the 2D image space, the color $C(\bm{x})$ for each pixel $\bm{x}$ is rendered by first sorting the splatted Gaussians along the depth and then performing the volumetric $\alpha$-blending of $\mathcal{N}$ ordered Gaussians as 
\begin{align}
    C(\bm{x}) = \sum_{i \in \mathcal{N}} \bm{c}_i \alpha_i \prod_{j=1}^{i-1}(1 - \alpha_j), ~~ \alpha_i = o_i \mathcal{G}_i^\text{S}(\bm{x})
\end{align}
where $\mathcal{G}_i^\text{S}(\bm{x})$ denotes the splatted $i^\text{th}$ Gaussian evaluated at the pixel $\bm{x}$.

\begin{figure}[!t]
    \centering
    \includegraphics[width=0.48\textwidth]{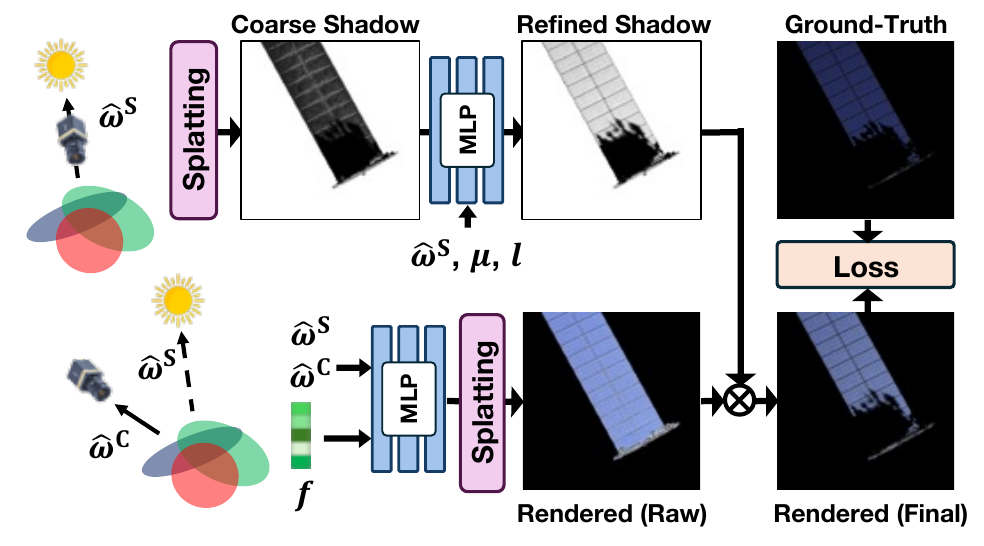}
    \caption{Proposed 3DGS pipeline with shadow splatting using the known Sun vector.}
    \label{fig:pipeline}
\end{figure}

\mysubsection{Leveraging the Sun Vector}
As noted in Sec.~\ref{sec:intro}, conventional in-the-wild methods are incompatible with spaceborne RPO scenarios due to constrained on-board computational resources which limit test-time optimization via differentiable rasterization of per-image appearance embeddings. Therefore, this work proposes to leverage the known Sun vector maintained by the servicer's AOCS.

The overall pipeline is illustrated in Fig.~\ref{fig:pipeline}. The core of the pipeline is the rasterization of appearance-conditioned Gaussians. Specifically, instead of directly learning the view-dependent color as done in the original 3DGS, each $i^\text{th}$ Gaussian is equipped with a trainable feature vector $\bm{f}_i \in \mathbb{R}^{72}$ that captures varying colors of Gaussians under different environments. In conventional in-the-wild methods such as WildGaussians \cite{kulhanek_2024_nips_wildgaussians}, each $j^\text{th}$ image is assigned a learnable embedding vector $\bm{e}_j$, so that the view-dependent color (i.e., SH coefficients) of that Gaussian is obtained by forward propagation of an appearance MLP $\bm{\Phi}$ (3 $\times$ 256-D hidden layers) given as $\bm{c}_i = \bm{\Phi}(\bm{f}_i, \bm{e}_j)$.

The proposed pipeline instead replaces the per-image embedding $\bm{e}_j$ with the Sun vector $\hat{\bm{\omega}}_j^\text{S}$ and the camera viewing vector $\hat{\bm{\omega}}_{ij}^\text{C}$ from the $i^\text{th}$ Gaussian. Following the practice of NeRF \cite{mildenhall_2020_eccv_nerf}, these unit vectors are first passed through the positional encoding function $\gamma_L(\cdot)$ defined as
\begin{equation} \label{eqn:position_enc}
    \begin{aligned}
        \gamma_L(p) = [ ~ \sin(2^0 \pi p) ~ \cos(2^0 \pi p) ~ \sin(2^1 \pi p) ~ \cos(2^1 \pi p) ~ \cdots \\
            ~ \sin(2^{L-1} \pi p) ~ \cos(2^{L-1} \pi p) ~ ]
    \end{aligned}  
\end{equation}
Then, the final view-dependent color of the $i^\text{th}$ Gaussian given known Sun and view vectors are obtained via
\begin{align}
    \bm{c}_i = \bm{\Phi}\big(\bm{f}_i, \gamma(\hat{\bm{\omega}}_j^\text{S}), \gamma(\hat{\bm{\omega}}_{ij}^\text{C}) \big)
\end{align}
and used in the 3DGS rasterization process.

The final rendered image $\hat{\bm{C}}_j$ and the observed ground-truth image $\bm{C}_j$ are used to compute the classical loss function for 3DGS
\begin{align} \label{eqn:loss}
    \mathcal{L}_\text{3DGS} = (1 - \lambda_\text{SSIM})\mathcal{L}_\text{1} + \lambda_\text{SSIM} \mathcal{L}_\text{SSIM}
\end{align}
where $\mathcal{L}_\text{1}$ is the mean pixel-wise $\ell_1$-distance, $\mathcal{L}_\text{SSIM}$ computes the Structural Similarity Index Measure (SSIM) \cite{wang_2004_tip_ssim}, and $\lambda_\text{SSIM} = 0.2$ is the weight hyperparameter. Inspired by GS-SLAM \cite{matsuki_2024_cvpr_gsslam}, an isotropic scale regularization is also considered, defined as
\begin{align} \label{eqn:isotropic}
    \mathcal{L}_\text{iso} = \sum_{i =1}^{|\mathcal{G}|} \| \bm{s}_i - \tilde{s}_i \cdot \bm{1} \|_1
\end{align}
where $\bm{s}_i$ is the scale parameters of the $i^\text{th}$ 3D Gaussian, and $\tilde{s}_i$ is its mean value. $\mathcal{L}_\text{iso}$ encourages sphericality and discourages elongated or otherwise irregular Gaussian shapes. While the original intention was to facilitate continuous tracking in SLAM, it has an added benefit when coupled with the shadow splatting---by discouraging lopsided geometries, the rendered images result in crisper and sharper shadows as expected in spaceborne imagery.

The final training loss is the combination of Eqs.~\ref{eqn:loss} and \ref{eqn:isotropic}
\begin{align} \label{eqn:final_loss}
    \mathcal{L} = \mathcal{L}_\text{3DGS} + \lambda_\text{iso} \mathcal{L}_\text{iso}
\end{align}

\mysubsection{Shadow Splatting}
In space, directional sunlight often casts extreme shadows on RSO due to self-occlusion. Such a phenomenon can be difficult for an MLP to learn without any regularization. Therefore, the proposed pipeline adopts the shadow splatting mechanism of GS${}^3$ \cite{bi_2024_siggraph_gs3} in order to inform each Gaussian whether it is lit or shadowed given the Sun's relative position. This is cleverly done by reusing the existing 3DGS rasterization framework. As visualized in Fig.~\ref{fig:pipeline}, the Gaussians are first splatted towards the light source for the $j^\text{th}$ image, which is equivalent to placing the camera center at the source with its boresight aligned with $-\hat{\bm{\omega}}_j^\text{S}$. Considering the distance to the Sun, the camera is instead ``placed'' at the original distance but along the Sun vector. Once splatted, the cumulative opacity of the Gaussians sorted along the ray is used to determine the visibility (i.e., shadow) of each $i^\text{th}$ Gaussian ($V_i$). The shadow value of each Gaussian is further refined with a small MLP $\bm{\Psi}$ (3 $\times$ 32-D hidden layers) via
\begin{align}
    V_i^\prime = \bm{\Psi}( V_i, \gamma(\hat{\bm{\omega}}_j^\text{S}) ~|~ \gamma(\bm{\mu}_i), \bm{l}_i )
\end{align}
where $\bm{\mu}_i$ is the $i^\text{th}$ Gaussian center location in the object frame and $\bm{l}_i \in \mathbb{R}^6$ is a trainable shadow latent vector.

The refined shadow values ($\bm{V}^\prime$) are finally splatted onto a shadow image and pixel-wise multiplied to the rendered image. The resulting image is used to compute the loss function in Eq.~\ref{eqn:loss}.

\mysubsection{Keyframe Management Strategy}
Given the limited on-board processing capability and staying in tune with previous works \cite{matsuki_2024_cvpr_gsslam, barad_2024_isparo_3dgs}, a small window of maximum $\mathcal{W}$ keyframes (i.e., images) is maintained. Existing SLAM methods select keyframes that avoid unnecessary redundancy while achieving significant baselines between subsequent keyframes with high covisibility of the shared observations \cite{murartal_2015_robotics_orbslam}. However, considering the scope of this paper and the assumption of known pose labels, the proposed 3DGS pipeline adopts a simple heuristic analogous to that of DSO \cite{engel_2018_tpami_dso} for keyframe management when training on a sequence of images collected during RPO. Let the window consist of keyframes $\mathcal{I}_1, \mathcal{I}_2, \dots, \mathcal{I}_n$ ordered in a reverse chronological order.
\begin{enumerate}
    \item The next frame $\mathcal{I}$ is added if its camera view vector $\hat{\bm{\omega}}^\text{C}$ is more than $\theta_\text{VIEW} = 10^\circ$ off from that of the last frame $\mathcal{I}_1$, i.e.,
    \begin{align}
        \hat{\bm{\omega}}^\text{C} \cdot \hat{\bm{\omega}}_1^\text{C} < \cos \theta_\text{VIEW}
    \end{align}
    \item The last two frames $\mathcal{I}_{1}, \mathcal{I}_{2}$ are always kept.
    \item If $n > \mathcal{W}$, then the oldest keyframe ($\mathcal{I}_i$) that has the smallest angular distance to its nearest neighbor is removed in order to promote a more "spread-out" set of keyframes in terms of the camera view vector, i.e., the one that maximizes the following distance score:
    \begin{align}
        d(\bm{\mathcal{I}}_i) = \max_{j \in [3, n] \setminus \{i\}} \hat{\bm{\omega}}_i^\text{C} \cdot \hat{\bm{\omega}}_j^\text{C}
    \end{align}
\end{enumerate}

\section{Experiments} \label{sec:experiments}

\begin{figure}[!t]
\centering
    \centering
    \includegraphics[width=0.45\textwidth]{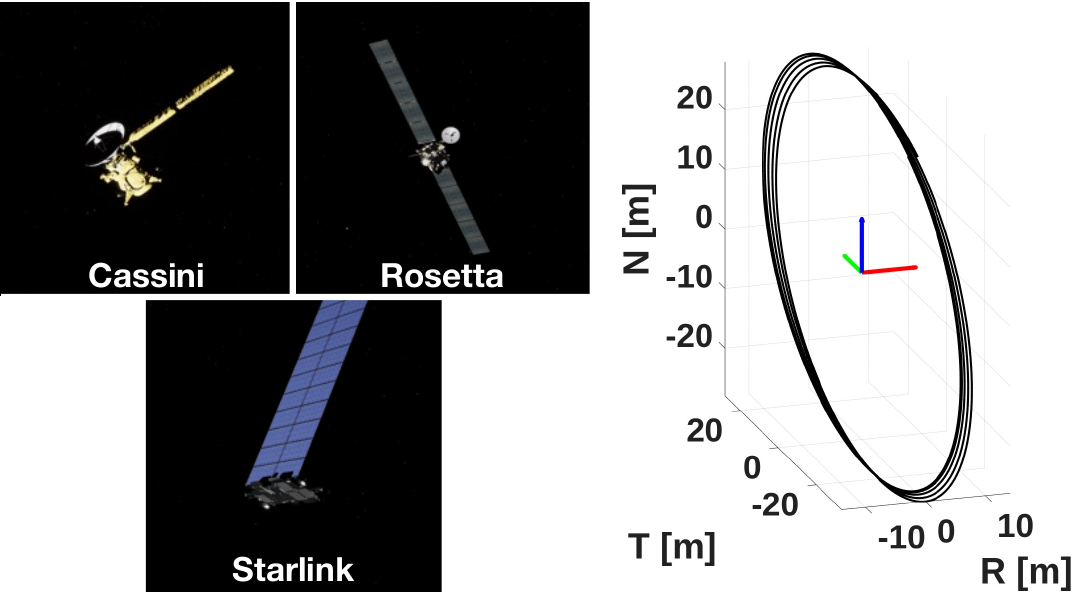}
    \caption{(\emph{Left}) Visualization of the adopted satellite models. (\emph{Right}) The servicer's orbit relative to the target in the RTN frame.}
    \label{fig:satellites}
\end{figure}

\begin{table*}[!t]
\caption{Quantitative performances of different configurations trained on sequential datasets. Arrows indicate the direction towards better performance. Bold faces denote the best performances. $| \mathcal{G} |$ denotes the number of Gaussians.}
\label{tab:results:sequential}
\centering
\footnotesize
\tabcolsep=0.1cm 
\begin{tabular}{lcccccccccccc}
\toprule
\multirow{2}{*}{Config.} & \multicolumn{4}{c}{Cassini} & \multicolumn{4}{c}{Rosetta} & \multicolumn{4}{c}{Starlink} \\
\cmidrule(lr){2-5} \cmidrule(lr){6-9} \cmidrule(lr){10-13}
& SSIM$\uparrow$ & PSNR$\uparrow$ & LPIPS$\downarrow$ & $|\mathcal{G}|$ & SSIM$\uparrow$ & PSNR$\uparrow$ & LPIPS$\downarrow$ & $|\mathcal{G}|$ & SSIM$\uparrow$ & PSNR$\uparrow$ & LPIPS$\downarrow$ & $|\mathcal{G}|$ \\
\midrule
(a) 3DGS \cite{kerbl_2023_acm_3dgs} & 0.9871 & 26.51 & 0.0171 & 42.7K & 0.9844 & 29.61 & 0.0187 & 48.7K & 0.9539 & 25.05 & 0.0440 & 74.4K \\
(b) + Illum. & 0.9894 & 30.63 & 0.0152 & 51.6K & 0.9889 & 33.40 & 0.0131 & 53.9K & 0.9657 & 27.58 & 0.0299 & 88.5K \\
(c) + Shadow Splatting & 0.9914 & 32.86 & \bfseries0.0131 & 45.9K & 0.9897 & 34.53 & 0.0123 & 54.4K & \bfseries0.9705 & 29.06 & \bfseries0.0251 & 89.7K \\
(d) + Isotropic Loss & \bfseries0.9916 & \bfseries33.40 & 0.0136 & 49.7K & \bfseries0.9905 & \bfseries35.27 & \bfseries0.0109 & 61.8K & 0.9683 & \bfseries29.89 & 0.0289 & 99.6K \\
\bottomrule
\end{tabular}
\end{table*}

\begin{figure*}[!t]
\centering
    \centering
    \includegraphics[width=0.975\textwidth]{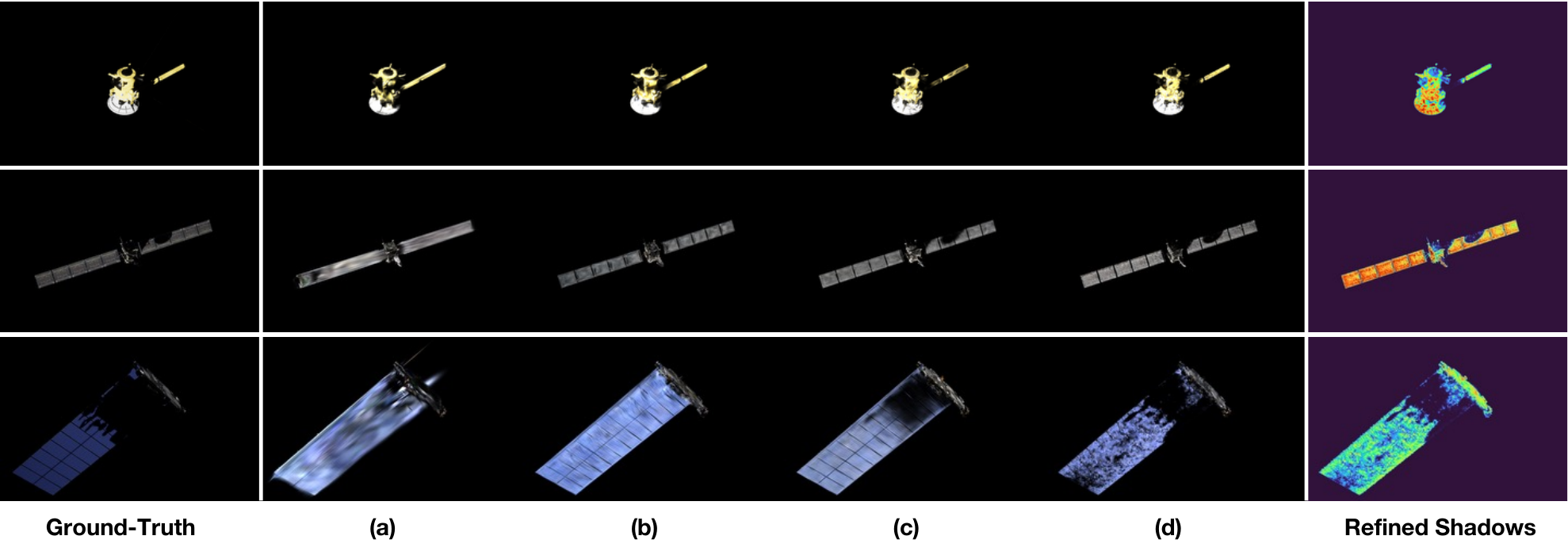}
    \caption{Qualitative results of different configurations trained on sequential datasets.}
    \label{fig:results:sequential}
\end{figure*}

\mysubsection{Datasets}
In order to train and validate the proposed pipeline, high-fidelity synthetic images are rendered using an Unreal Engine (UE)-based simulator \cite{park_2024_scitech_spe3r} for three different satellite models with varying geometric features as shown in Fig.~\ref{fig:satellites}. Specifically, Cassini and Rosetta 3D models are obtained from the ESA Science Satellite Fleet \cite{esascifleet}, while the Starlink 3D model is CC-licensed and available online \cite{starlink_3dmodel}. All images are rendered using the camera intrinsics adopted in the SPEED \cite{sharma_2020_taes_spn} dataset. Since SPEED camera intrinsics result in very high resolution images (1920 $\times$ 1200), all images are downsampled by the factor of 2 throughout the experiments.

For each satellite, a new dataset is created which comprises images captured during a simulated RPO trajectory in Low Earth Orbit (LEO) in which the servicer is in a passively safe $e$/$i$-vector-separated relative orbit \cite{montenbruck_2006_ast_eivectorsep} about the target RSO. The servicer's camera is always pointed at the target, and the images are captured every 5 seconds for 5 orbits, which amounts to 5,764 images per simulation. In order to obtain diverse views of the target more frequently, the target is simulated to tumble about its $\hat{x}$-axis at 2${}^\circ$/sec unless noted otherwise. The servicer's relative orbit is visualized in Fig.~\ref{fig:satellites}.

Finally, in order to evaluate the generalizability of the trained 3DGS models, they are evaluated on 100 images of each satellite rendered with random illumination conditions and camera and target poses.

\mysubsection{Implementation Details}
The training begins once the keyframe window is full. Specifically, a round of optimization is performed on a maximum $\mathcal{W}$ = 10 keyframes (i.e., one round is equivalent to 10 global iterations) unless noted otherwise. The optimization is triggered only when the composition of the keyframe window changes. The Gaussians are added and pruned according to the original 3DGS implementation once every 10 rounds. Since the number of training rounds could vary in RPO settings, the learning rates are repeatedly decayed within each training round, resulting in a repeated sawtooth-like pattern throughout the training. The underlying implementation and all other training parameters follow the original 3DGS repository\footnote{\url{https://github.com/graphdeco-inria/gaussian-splatting}}. Recall that the Gaussian positions are initialized by uniformly sampling 10,000 points from the 3D model surfaces.

Four different architectural configurations are compared during the experiments. (a) The original 3DGS \cite{kerbl_2023_acm_3dgs} learns per-Gaussian color directly from the differentiable rasterization and serves as the baseline. (b) Per-Gaussian appearance features ($\bm{f}$) and the known Sun and camera view vectors ($\hat{\bm{\omega}}^\text{S}$, $\hat{\bm{\omega}}^\text{C}$) are added and optimized via an appearance MLP ($\Phi$). (c) The shadow splatting pipeline is added, and the resulting refined shadow image ($\bm{V}^\prime$) is multiplied to the final rendered image from the main Gaussian splatting pipeline. (d) Finally, the isotropic scale regularization term (Eq.~\ref{eqn:isotropic}) is added to the final loss function.

\mysubsection{Metrics}
This paper adopts three standard quantitative metrics commonly used in radiance fields literature: SSIM \cite{wang_2004_tip_ssim}, Peak Signal-to-Noise Ratio (PSNR), and Learned Perceptual Image Patch Similarity (LPIPS) \cite{zhang_2018_cvpr_lpips}. Since the initial Gaussians are sampled from the ground-truth 3D model, the geometric accuracy of reconstructed 3DGS models is not compared with the ground-truth models using 3D metrics such as Chamfer distance \cite{park_2024_scitech_spe3r}.

\section{Results}

The qualitative and quantitative performances of four different model configurations introduced in Sec.~\ref{sec:experiments} are respectively shown in Fig.~\ref{fig:results:sequential} and Table \ref{tab:results:sequential}. It is immediately clear that the original 3DGS struggles with learning from dynamic illumination conditions in the synthetic images. Adding an appearance MLP and providing the Sun vector significantly improves performance across all metrics. However, Figure \ref{fig:results:sequential} shows that the reconstructed model fails to capture the global illumination and self-occlusion even with perfect knowledge of the Sun's position. Once the shadow splatting is incorporated into the pipeline, the performance once again visibly improves across all metrics, and the reconstructed model is capable of capturing the shadows caused by self-occlusion, albeit blurry and approximate. Finally, Figure \ref{fig:results:sequential} demonstrates that including the isotropic scale regularization ($\mathcal{L}_\text{iso}$) term into the final loss function enables rendering of sharp shadows, but the overall reconstruction quality, especially that of the Starlink satellite with a massive solar panel, degrades a little as evidenced by drops in SSIM and LPIPS metrics as reported in Table \ref{tab:results:sequential}.

\begin{figure}[!t]
\centering
    \centering
    \includegraphics[width=0.48\textwidth]{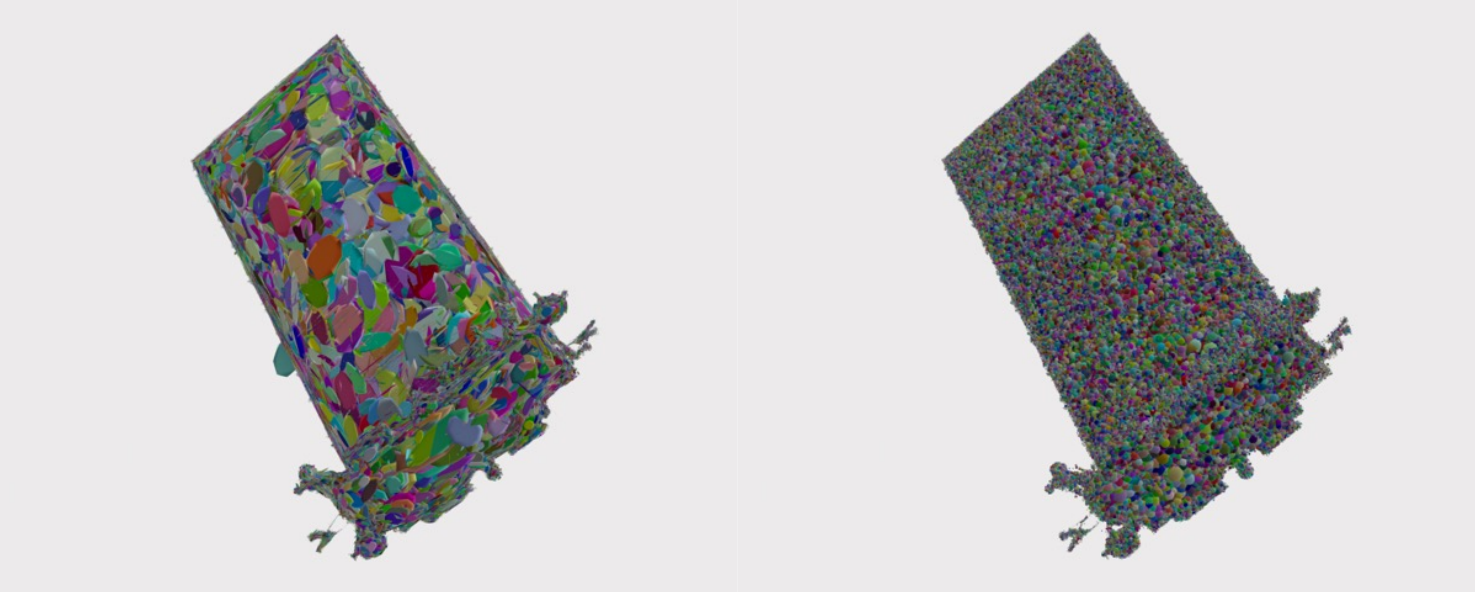}
    \caption{Visualization of raw Gaussians after training without (\emph{left}) and with (\emph{right}) the $\mathcal{L}_\text{iso}$ term. }
    \label{fig:results:isotropic_naked}
\end{figure}

The effect of $\mathcal{L}_\text{iso}$ is illustrated in Fig.~\ref{fig:results:isotropic_naked} where the 3D Gaussians are visualized without 2D splatting but instead with random colors. It can be seen that training with the $\mathcal{L}_\text{iso}$ term causes the learned Gaussians to become much more spherical and granular, discouraging large and elongated shapes often encountered in the model trained without the regularization. Smaller Gaussians enable capturing sharper shadow contours since the effect of direct sunlight becomes more localized. However, it also has the side-effect of an increased number of Gaussians as shown in Table \ref{tab:results:sequential}.

\begin{figure}[!t]
\centering
    \centering
    \includegraphics[width=0.48\textwidth]{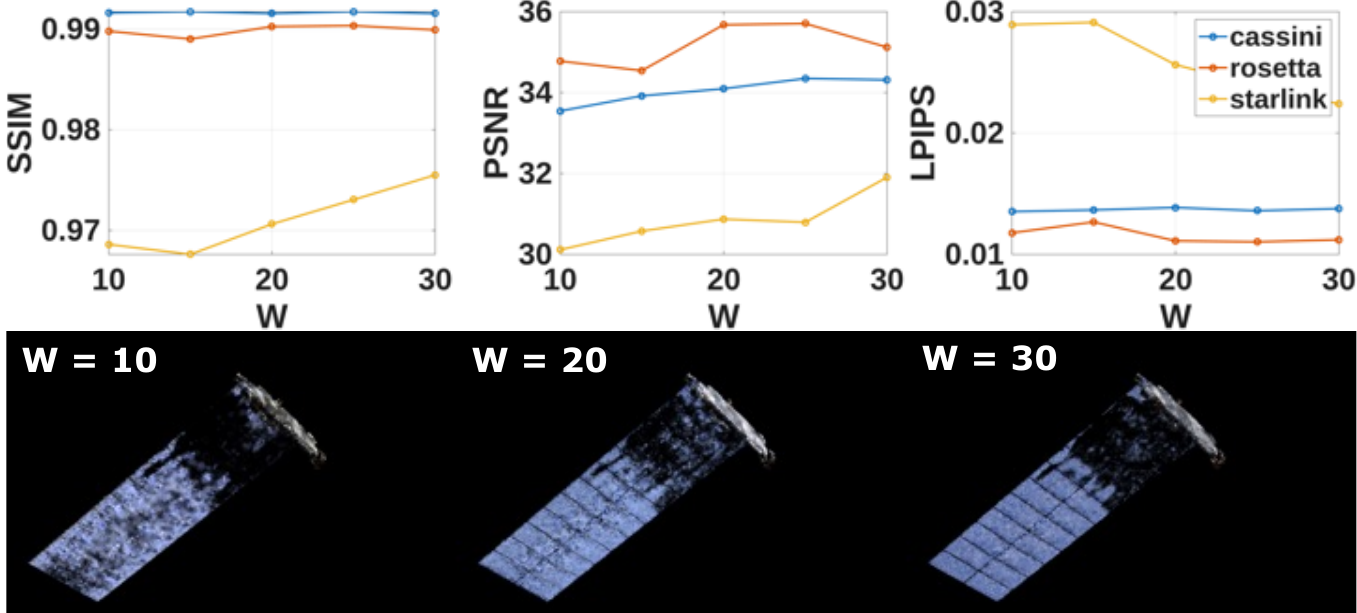}
    \caption{Performance as the keyframe window size ($\mathcal{W}$) varies}
    \label{fig:results:ablation_window}
\end{figure}

Finally, a small ablation study is conducted by varying the length of the keyframe window ($\mathcal{W}$) using the full model configuration. Figure \ref{fig:results:ablation_window} reports that increasing $\mathcal{W}$ improves performance across all metrics, and the improvement is most distinct for the Starlink satellite. For instance, when the window length is as small as $\mathcal{W} = 10$, the reconstructed quality of the Starlink's solar panel is poor with many Gaussians unable to reconstruct proper colors. This is an expected behavior since raising $\mathcal{W}$ not only increases the overall number of training steps but also improves the diversity of samples that are observed at each training round. However, there is a trade-off in varying the window size, i.e., the number of images and associated metadata to maintain, and the on-board computational and memory capacity of satellites.

\begin{figure}[!t]
    \centering
    \includegraphics[width=0.48\textwidth]{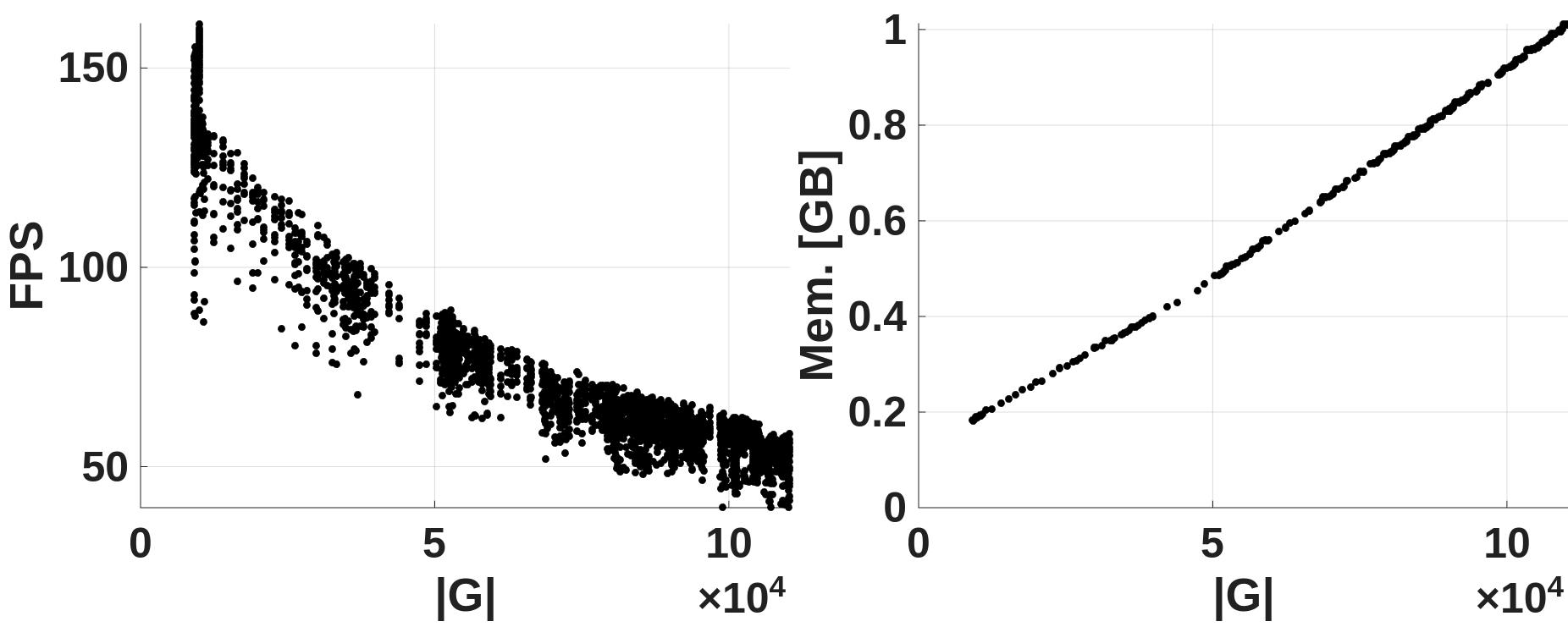}
    \caption{Training speed and incurred memory vs.~number of Gaussians ($|\mathcal{G}|$) measured on an NVIDIA GeForce RTX 4090 GPU.}
    \label{fig:results:efficiency}
\end{figure}

In fact, Fig.~\ref{fig:results:efficiency} presents a preliminary analysis of the training speed and GPU memory usage during a single training step. Each training step comprises both forward and backward propagation through the full rasterization pipeline shown in Fig.~\ref{fig:pipeline} including shadow splatting for a single 960 $\times$ 600 image input. The reported GPU memory denotes the amount allocated by the aforementioned training step\footnote{The memory is computed as the difference of the PyTorch's \texttt{torch.cuda.max\_memory\_allocated()} calls before and after the training step.} which excludes the memory reserved for the training data. 

At nearly 100,000 Gaussians, the proposed pipeline still trains at about 50 frames per second on a desktop GPU while incurring an additional 1 GB of GPU memory. When more images are kept in the keyframe window, the additional memory overhead will inevitably increase which requires careful analysis. Meanwhile, the training speed can be improved by maintaining an optimal number of Gaussians throughout training, which in turn demands a more robust densification strategy.

\begin{figure}[!t]
    \centering
    \includegraphics[width=0.48\textwidth]{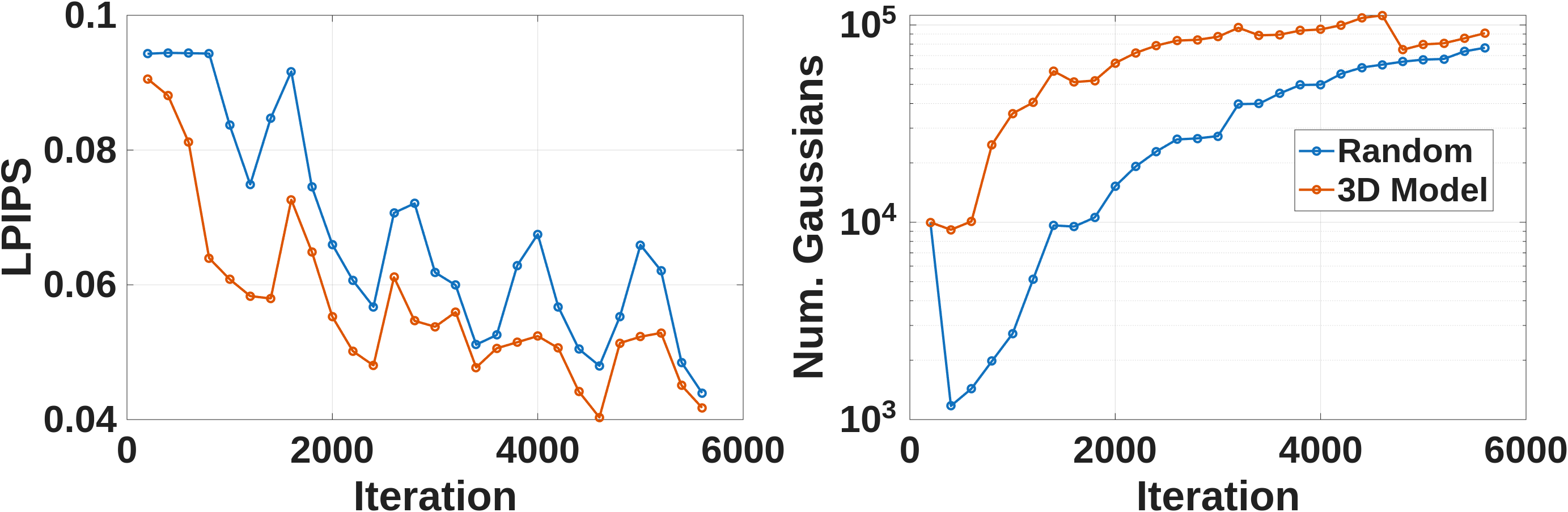}
    \caption{Comparison of the LPIPS metric and the number of Gaussians when the model is initialized as 10,000 points sampled randomly and from the ground-truth 3D model.}
    \label{fig:initialization}
\end{figure}

\section{Discussions} \label{sec:discussions}
As noted at the beginning of Sec.~\ref{sec:method}, this work has made a number of ideal assumptions in order to focus on the photometric accuracy of the trained 3DGS model under dynamic illumination conditions in space. This section discusses how these assumptions could be relaxed moving forward.

\mysubsection{Initialization of Gaussians} 
The conventional initialization method via SfM is not compatible with RPO missions, as SfM algorithms struggle under changing illumination conditions and require a large batch of images. However, as portrayed in Fig.~\ref{fig:initialization}, proper initialization of the Gaussians can significantly accelerate the model convergence. While the model can still be trained from a completely random set of points, it initially suffers from a huge loss of Gaussians as the densification strategy kicks in, further slowing down the training.

Several remedies are being investigated and adopted in the literature. One is to leverage zero-shot shape abstraction algorithms, such as that of Park \& D'Amico \cite{park_2024_scitech_spe3r}, which predict coarse, normalized 3D structures of the target and the camera poses simultaneously from single images. For instance, the concurrent work by Francesch Huc et al.~\cite{francesch_2025_asc_sq3dgs} successfully initializes 3DGS models using point clouds sampled from the surfaces of reconstructed superquadric assemblies. Another potential solution is to employ additional sensor modalities such as depth cameras that can be used to initialize points along the visible surfaces \cite{barad_2024_isparo_3dgs}.

\mysubsection{Availability of Foreground Masks} 
Foreground masks allow background removal which helps concentrate Gaussians to the target RSO in question. Depth sensors can be used in this case as well to swiftly isolate the nearby objects from far-away backgrounds (e.g., Earth). One can also utilize general-purpose segmentation models (e.g., Segment Anything \cite{kirillov_2023_segmentanything}) fine-tuned specifically for spaceborne missions.

\mysubsection{Knowledge of Pose Labels} 
In real missions, pose labels are not available and must be jointly estimated as the 3DGS model is learned. This can be achieved through SLAM-like procedures \cite{murartal_2015_robotics_orbslam, matsuki_2024_cvpr_gsslam, barad_2024_isparo_3dgs}, in which case the keyframe management system must rely on the covisibility of Gaussians rather than noisy pose estimates. Another possibility is to directly integrate the pose predictions from the aforementioned zero-shot method \cite{park_2024_scitech_spe3r, bates_2025_gnc_ambiguity} into the SLAM pipeline.

\section{Conclusions}

This paper presented a novel pipeline to train a 3D Gaussian Splatting (3DGS) model using sequences of target Resident Space Object (RSO) images captured during Rendezvous and Proximity Operations (RPO) in space. Specifically, considering extreme and dynamic illumination conditions in spaceborne imagery and the fact that 3DGS assumes static scenes, the proposed pipeline incorporates (1) the Sun vector tracked by the servicer's onboard sensors and (2) shadow splatting, which leverages an existing splatting algorithm, to evaluate the visibility of each Gaussian given the known lighting direction at each image. The experimental results on high-fidelity synthetic images captured during the simulations of representative RPO scenarios indicate that the proposed method is capable of learning a 3DGS model of the target RSO from an image sequence, able to capture global effects of changing Sun directions and self-occlusion. 

The proposed pipeline currently hinges on a number of ideal assumptions, such as known pose labels and initialization of Gaussians from a ground-truth 3D mesh model. Future plans to relax these assumptions are discussed, including the adoption of additional sensor modalities and SLAM-like mapping and tracking procedures. 

\section*{Acknowledgment}
The authors would like to thank Dr.~Kuldeep Barad at Redwire Space Europe and Emily Bates and Pol Francesch Huc at SLAB for helpful discussions and constructive feedback.

\bibliographystyle{IEEEtran}
\bibliography{reference}

@IEEEtranBSTCTL{IEEEexample:BSTcontrol,
    CTLuse_forced_etal = "yes",
    CTLmax_names_forced_etal = "3",
    CTLnames_show_etal = "3",
}

@inproceedings{kulhanek_2024_nips_wildgaussians,
 author = {Kulhanek, Jonas and Peng, Songyou and Kukelova, Zuzana and Pollefeys, Marc and Sattler, Torsten},
 booktitle = {Advances in Neural Information Processing Systems},
 pages = {21271--21288},
 title = {{WildGaussians}: {3D} Gaussian Splatting In the Wild},
 year = {2024}
}

@INPROCEEDINGS{martinbrualla_2021_cvpr_nerfw,
  author={Martin-Brualla, Ricardo and Radwan, Noha and Sajjadi, Mehdi S. M. and Barron, Jonathan T. and Dosovitskiy, Alexey and Duckworth, Daniel},
  booktitle={2021 IEEE/CVF Conference on Computer Vision and Pattern Recognition (CVPR)}, 
  title={{NeRF in the Wild}: Neural Radiance Fields for Unconstrained Photo Collections}, 
  year={2021},
  volume={},
  number={},
  pages={7206-7215},
  doi={10.1109/CVPR46437.2021.00713}
}

@article{kerbl_2023_acm_3dgs,
    author = {Kerbl, Bernhard and Kopanas, Georgios and Leimkuehler, Thomas and Drettakis, George},
    title = {{3D} Gaussian Splatting for Real-Time Radiance Field Rendering},
    year = {2023},
    issue_date = {August 2023},
    publisher = {Association for Computing Machinery},
    address = {New York, NY, USA},
    volume = {42},
    number = {4},
    doi = {10.1145/3592433},
    journal = {ACM Trans. Graph.},
    month = jul,
    articleno = {139},
    numpages = {14},
}

@INPROCEEDINGS{zwicker_2001_vis_ewasplatting,
  author={Zwicker, M. and Pfister, H. and van Baar, J. and Gross, M.},
  booktitle={Proceedings Visualization, 2001. VIS '01.}, 
  title={{EWA} volume splatting}, 
  year={2001},
  volume={},
  number={},
  pages={29-538},
  doi={10.1109/VISUAL.2001.964490}
}

@inproceedings{pyrak_2022_amos_mev,
    author = {Matt Pyrak and Joseph Anderson},
    title = {{Performance of Northrop Grumman’s Mission Extension Vehicle (MEV) RPO imagers at GEO}},
    booktitle = {Autonomous Systems: Sensors, Processing and Security for Ground, Air, Sea and Space Vehicles and Infrastructure 2022},
    pages = {121150A},
    year = {2022},
    doi = {10.1117/12.2631524},
}

@article{reed_aiaaspace_2016_restorel, 
    title={{The Restore-L Servicing Mission}}, 
    DOI={10.2514/6.2016-5478}, 
    journal={AIAA Space 2016}, 
    author={Reed, Benjamin B. and Smith, Robert C. and Naasz, Bo J. and Pellegrino, Joseph F. and Bacon, Charles E.}, 
    year={2016}
}

@article{aglietti_2019_aer_removedebris,
    title={{RemoveDEBRIS}: An in-orbit demonstration of technologies for the removal of space debris},
    volume={124},
    DOI={10.1017/aer.2019.136},
    number={1271},
    journal={The Aeronautical Journal},
    publisher={Cambridge University Press},
    author={Aglietti, G. S. and Taylor, B. and Fellowes, S. and Ainley, S. and Tye, D. and Cox, C. and Zarkesh, A. and Mafficini, A. and Vinkoff, N. and Bashford, K. and Salmon, T. and Retat, I. and Burgess, C. and Hall, A. and Chabot, T. and Kanani, K. and Pisseloup, A. and Bernal, C. and Chaumette, F. and Pollini, A. and Steyn, W. H.},
    year={2020},
    pages={1–23}
}

@inproceedings{park_2019_aas_krn,
	author={Park, Tae Ha and Sharma, Sumant and D'Amico, Simone}, 
	booktitle={2019 AAS/AIAA Astrodynamics Specialist Conference, Portland, Maine}, 
	title={Towards Robust Learning-Based Pose Estimation of Noncooperative Spacecraft}, 
	year={2019}, 
	volume={}, 
	number={}, 
	pages={}, 
	month={August 11-15}
}

@article{park_2024_asr_spnv2,
  title = {Robust multi-task learning and online refinement for spacecraft pose estimation across domain gap},
  author = {Tae Ha Park and Simone D’Amico},
  journal = {Advances in Space Research},
  volume = {73},
  number = {11},
  pages = {5726--5740},
  year = {2024},
  note = {Recent Advances in Satellite Constellations and Formation Flying},
  doi = {10.1016/j.asr.2023.03.036},
}

@INPROCEEDINGS{park_2024_icra_ost,
  author={Park, Tae Ha and D’Amico, Simone},
  booktitle={2024 IEEE International Conference on Robotics and Automation (ICRA)},
  title={Online Supervised Training of Spaceborne Vision during Proximity Operations using Adaptive Kalman Filtering},
  year={2024},
  volume={},
  number={},
  pages={11744-11752},
  doi={10.1109/ICRA57147.2024.10610138}
}

@ARTICLE{sharma_2020_taes_spn,
	author={Sharma, Sumant and D’Amico, Simone},
	journal={IEEE Transactions on Aerospace and Electronic Systems}, 
	title={Neural Network-Based Pose Estimation for Noncooperative Spacecraft Rendezvous}, 
	year={2020},
	volume={56},
	number={6},
	pages={4638-4658},
	doi={10.1109/TAES.2020.2999148}
}

@INPROCEEDINGS{sharma_2018_aero_pose,
  author={Sharma, Sumant and Beierle, Connor and D'Amico, Simone},
  booktitle={2018 IEEE Aerospace Conference}, 
  title={Pose estimation for non-cooperative spacecraft rendezvous using convolutional neural networks}, 
  year={2018},
  volume={},
  number={},
  pages={1-12},
  doi={10.1109/AERO.2018.8396425}
}

@article{damico_2014_ijsse_pose,
    title={Pose estimation of an uncooperative spacecraft from actual space imagery},
    volume={2},
    DOI={10.1504/ijspacese.2014.060600},
    number={2},
    journal={International Journal of Space Science and Engineering},
    author={D'Amico, Simone and Benn, Mathias and J{\o}rgensen, John L.},
    year={2014},
    pages={171}
}

@INPROCEEDINGS{chen_2019_iccvw_pose,
  author={Chen, Bo and Cao, Jiewei and Parra, Alvaro and Chin, Tat-Jun},
  booktitle={2019 IEEE/CVF International Conference on Computer Vision Workshop (ICCVW)}, 
  title={Satellite Pose Estimation with Deep Landmark Regression and Nonlinear Pose Refinement}, 
  year={2019},
  volume={},
  number={},
  pages={2816-2824},
  doi={10.1109/ICCVW.2019.00343}
}

@INPROCEEDINGS{proenca_2020_icra_urso,
    author={Proen\c{c}a, Pedro F. and Gao, Yang},
    booktitle={2020 IEEE International Conference on Robotics and Automation (ICRA)}, 
    title={Deep Learning for Spacecraft Pose Estimation from Photorealistic Rendering}, 
    year={2020},
    volume={},
    number={},
    pages={6007-6013},
    doi={10.1109/ICRA40945.2020.9197244}
}

@article{sharma_2018_jsr_robust, 
	title={Robust Model-Based Monocular Pose Initialization for Noncooperative Spacecraft Rendezvous}, 
	author={Sharma, Sumant and Ventura, Jacopo and D’Amico, Simone},
	DOI={10.2514/1.a34124}, 
	journal={Journal of Spacecraft and Rockets}, 
	volume={55},
    number={6},
    pages={1414--1429},
    year={2018},
}

@inproceedings{capuano_2019_scitech_robust,
	author = {Vincenzo Capuano and Shahrouz Ryan Alimo and Andrew Q. Ho and Soon-Jo Chung},
	title = {Robust Features Extraction for On-board Monocular-based Spacecraft Pose Acquisition},
	booktitle = {AIAA Scitech 2019 Forum},
	chapter = {},
	pages = {},
    year={2019},
	doi = {10.2514/6.2019-2005}
}

@misc{grompone_2015_phd,
  title = {Vision-based {3D} motion estimation for on-orbit proximity satellite tracking and navigation},
  author = {Grompone, Alessio A.},
  year = {2015},
  URL = {https://calhoun.nps.edu/handle/10945/45863},
  publisher = {Calhoun}
}

@INPROCEEDINGS{petit_2011_iros_pose,
  author={Petit, Antoine and Marchand, Eric and Kanani, Keyvan},
  booktitle={2011 IEEE/RSJ International Conference on Intelligent Robots and Systems}, 
  title={Vision-based space autonomous rendezvous: A case study}, 
  year={2011},
  volume={},
  number={},
  pages={619-624},
  doi={10.1109/IROS.2011.6094568}
}

@article{park_2023_jgcd_spnukf,
	title={Adaptive Neural-Network-Based Unscented Kalman Filter for Robust Pose Tracking of Noncooperative Spacecraft},
	author={Park, Tae Ha and D'Amico, Simone},
	journal={Journal of Guidance, Control, and Dynamics},
	volume={46},
	number={9},
	pages={1671--1688},
	year={2023},
	doi={10.2514/1.G007387}
}

@article{pasqualetto_2021_acta_cnn,
    title = {Evaluation of tightly- and loosely-coupled approaches in {CNN}-based pose estimation systems for uncooperative spacecraft},
    journal = {Acta Astronautica},
    volume = {182},
    pages = {189-202},
    year = {2021},
    doi = {10.1016/j.actaastro.2021.01.035},
    author = {Lorenzo {Pasqualetto Cassinis} and Robert Fonod and Eberhard Gill and Ingo Ahrns and Jesús Gil-Fernández},
}

@article{pasqualetto_2023_acta_ukf,
    title = {Leveraging neural network uncertainty in adaptive unscented {Kalman} Filter for spacecraft pose estimation},
    author = {Pasqualetto Cassinis, Lorenzo and Park, Tae Ha and Stacey, Nathan and D’Amico, Simone and Menicucci, Alessandra and Gill, Eberhard and Ahrns, Ingo and Sanchez-Gestido, Manuel},
    journal = {Advances in Space Research},
    volume = {71},
    number = {12},
    pages = {5061-5082},
    year = {2023},
    issn = {0273-1177},
    doi = {10.1016/j.asr.2023.02.021},
}

@article{kisantal_2020_taes_spec,
    author={Kisantal, Mate and 
            Sharma, Sumant and
            Park, Tae Ha and
            Izzo, Dario and
            M\"{a}rtens, Marcus and
            D'Amico, Simone},
    journal={IEEE Transactions on Aerospace and Electronic Systems}, 
    title={Satellite Pose Estimation Challenge: Dataset, Competition Design and Results}, 
    year={2020},
    volume={56},
    number={5},
    pages={4083-4098},
    doi={10.1109/TAES.2020.2989063}
}

@article{park_2023_acta_spec2021,
	title = {Satellite Pose Estimation Competition 2021: Results and Analyses},
	author = {Park, Tae Ha and M\"{a}rtens, Marcus and Jawaid, Mohsi and Wang, Zi and Chen, Bo and Chin, Tat-Jun and Izzo, Dario and D’Amico, Simone},
	journal = {Acta Astronautica},
	volume = {204},
	pages = {640-665},
	year = {2023},
	doi = {10.1016/j.actaastro.2023.01.002},
}

@article{kaki_2023_jais_pose,
    author = {Kaki, Siddarth and Deutsch, Jacob and Black, Kevin and Cura-Portillo, Asher and Jones, Brandon A. and Akella, Maruthi R.},
    title = {Real-Time Image-Based Relative Pose Estimation and Filtering for Spacecraft Applications},
    journal = {Journal of Aerospace Information Systems},
    volume = {20},
    number = {6},
    pages = {290-307},
    year = {2023},
    doi = {10.2514/1.I011196},
}

@misc{park_2024_spnv3,
  title={Bridging the Domain Gap for Flight-Ready Spaceborne Vision}, 
  author={Tae Ha Park and Simone D'Amico},
  year={2024},
  eprint={2409.11661},
  archivePrefix={arXiv},
  primaryClass={cs.CV},
  url={https://arxiv.org/abs/2409.11661}, 
}

@INPROCEEDINGS{hu_2021_cvpr_swisscube,
  author={Hu, Yinlin and Speierer, S\'{e}bastien and Jakob, Wenzel and Fua, Pascal and Salzmann, Mathieu},
  booktitle={2021 IEEE/CVF Conference on Computer Vision and Pattern Recognition (CVPR)}, 
  title={Wide-Depth-Range {6D} Object Pose Estimation in Space}, 
  year={2021},
  volume={},
  number={},
  pages={15865-15874},
  doi={10.1109/CVPR46437.2021.01561}
}

@InProceedings{garcia_2021_cvprw_lspnet,
	author    = {Garcia, Albert and Musallam, Mohamed Adel and Gaudilliere, Vincent and Ghorbel, Enjie and Al Ismaeil, Kassem and Perez, Marcos and Aouada, Djamila},
	title     = {{LSPnet}: A 2D Localization-Oriented Spacecraft Pose Estimation Neural Network},
	booktitle = {Proceedings of the IEEE/CVF Conference on Computer Vision and Pattern Recognition (CVPR) Workshops},
	month     = {June},
	year      = {2021},
	pages     = {2048-2056}
}

@inproceedings{park_2022_aero_speedplus,
	author={Park, Tae Ha and M{\"a}rtens, Marcus and Lecuyer, Gurvan and Izzo, Dario and D'Amico, Simone},
	booktitle={2022 IEEE Aerospace Conference (AERO)},
	title={{SPEED+}: Next-Generation Dataset for Spacecraft Pose Estimation across Domain Gap},
	year={2022},
	pages={1-15},
	doi={10.1109/AERO53065.2022.9843439}
}

@INPROCEEDINGS{musallam_2021_icipc_spark,
	author={Musallam, Mohamed Adel and Gaudilliere, Vincent and Ghorbel, Enjie and Ismaeil, Kassem Al and Perez, Marcos Damian and Poucet, Michel and Aouada, Djamila},
	booktitle={2021 IEEE International Conference on Image Processing Challenges (ICIPC)}, 
	title={Spacecraft Recognition Leveraging Knowledge of Space Environment: Simulator, Dataset, Competition Design and Analysis}, 
	year={2021},
	volume={},
	number={},
	pages={11-15},
	doi={10.1109/ICIPC53495.2021.9620184}
}

@inproceedings{park_2021_aas_tron,
	author={Park, Tae Ha and Bosse, Juergen and D'Amico, Simone}, 
	booktitle={2021 AAS/AIAA Astrodynamics Specialist Conference, Big Sky, Vitrual}, 
	title={Robotic Testbed for Rendezvous and Optical Navigation: Multi-Source Calibration and Machine Learning Use Cases}, 
	year={2021}, 
	volume={}, 
	number={}, 
	pages={}, 
	month={August 9-11}
}

@article{pasqualetto_2022_acta_ogrl,
    title = {On-ground validation of a CNN-based monocular pose estimation system for uncooperative spacecraft: Bridging domain shift in rendezvous scenarios},
    journal = {Acta Astronautica},
    volume = {196},
    pages = {123-138},
    year = {2022},
    doi = {10.1016/j.actaastro.2022.04.002},
    author = {Lorenzo {Pasqualetto Cassinis} and Alessandra Menicucci and Eberhard Gill and Ingo Ahrns and Manuel Sanchez-Gestido},
}

@article{olivaresmendez_2023_jsse_zeroglab,
title = {{Zero-G Lab}: A multi-purpose facility for emulating space operations},
journal = {Journal of Space Safety Engineering},
volume = {10},
number = {4},
pages = {509-521},
year = {2023},
issn = {2468-8967},
doi = {10.1016/j.jsse.2023.09.003},
author = {Miguel Olivares-Mendez and Mohatashem Reyaz Makhdoomi and Barış Can Yalçın and Zhanna Bokal and Vivek Muralidharan and Miguel {Ortiz Del Castillo} and Vincent Gaudilliere and Leo Pauly and Olivia Borgue and Mohammadamin Alandihallaj and Jan Thoemel and Ernest Skrzypczyk and Arunkumar Rathinam and Kuldeep Rambhai Barad and Abd El Rahman Shabayek and Andreas M. Hein and Djamila Aouada and Carol Martinez},
}

@ARTICLE{murartal_2015_robotics_orbslam,
  author={Mur-Artal, Ra\'{u}l and Montiel, J. M. M. and Tard\'{o}s, Juan D.},
  journal={IEEE Transactions on Robotics}, 
  title={{ORB-SLAM}: A Versatile and Accurate Monocular {SLAM} System}, 
  year={2015},
  volume={31},
  number={5},
  pages={1147-1163},
  doi={10.1109/TRO.2015.2463671}
}

@inproceedings{dor_2018_sfm_orbslam,
    author = {Mehregan Dor and Panagiotis Tsiotras},
    title = {{ORB-SLAM} Applied to Spacecraft Non-Cooperative Rendezvous},
    booktitle = {2018 Space Flight Mechanics Meeting},
    chapter = {},
    pages = {},
    year={2018},
    doi = {10.2514/6.2018-1963},
}

@inproceedings{park_2024_scitech_spe3r,
	author = {Tae Ha Park and Simone D'Amico},
	title = {Rapid Abstraction of Spacecraft {3D} Structure from Single {2D} Image},
	booktitle = {AIAA SCITECH 2024 Forum},
	chapter = {},
	pages = {},
        year={2024},
	doi = {10.2514/6.2024-2768},
}

@InProceedings{mildenhall_2020_eccv_nerf,
    doi="10.1007/978-3-030-58452-8_24",
    author="Mildenhall, Ben
    and Srinivasan, Pratul P.
    and Tancik, Matthew
    and Barron, Jonathan T.
    and Ramamoorthi, Ravi
    and Ng, Ren",
    title="{NeRF}: Representing Scenes as Neural Radiance Fields for View Synthesis",
    booktitle="Computer Vision -- ECCV 2020",
    year="2020",
    pages="405--421"
}

@INPROCEEDINGS{rosinol_2023_iros_nerfslam,
  author={Rosinol, Antoni and Leonard, John J. and Carlone, Luca},
  booktitle={2023 IEEE/RSJ International Conference on Intelligent Robots and Systems (IROS)}, 
  title={{NeRF-SLAM}: Real-Time Dense Monocular {SLAM} with Neural Radiance Fields}, 
  year={2023},
  volume={},
  number={},
  pages={3437-3444},
  doi={10.1109/IROS55552.2023.10341922}}

@INPROCEEDINGS{matsuki_2024_cvpr_gsslam,
  author={Matsuki, Hidenobu and Murai, Riku and Kelly, Paul H. J. and Davison, Andrew J.},
  booktitle={2024 IEEE/CVF Conference on Computer Vision and Pattern Recognition (CVPR)}, 
  title={Gaussian Splatting {SLAM}}, 
  year={2024},
  volume={},
  number={},
  pages={18039-18048},
  doi={10.1109/CVPR52733.2024.01708}}

@PHDTHESIS {tweddle_2013_thesis,
    author  = "Tweddle, Brent Edward",
    title   = "Computer vision-based localization and mapping of an unknown, uncooperative and spinning target for spacecraft proximity operations",
    school  = "Massachusetts Institute of Technology",
    year    = "2013",
    address = "Department of Aeronautics and Astronautics",
    howpublished = "\url{https://dspace.mit.edu/handle/1721.1/85693}"
}

@article{bay_2008_surf,
    title = {Speeded-Up Robust Features ({SURF})},
    journal = {Computer Vision and Image Understanding},
    volume = {110},
    number = {3},
    pages = {346-359},
    year = {2008},
    doi = {10.1016/j.cviu.2007.09.014},
    author = {Herbert Bay and Andreas Ess and Tinne Tuytelaars and Luc {Van Gool}}
}

@ARTICLE{kaess_2008_robotics_isam,
  author={Kaess, Michael and Ranganathan, Ananth and Dellaert, Frank},
  journal={IEEE Transactions on Robotics}, 
  title={{iSAM}: Incremental Smoothing and Mapping}, 
  year={2008},
  volume={24},
  number={6},
  pages={1365-1378},
  doi={10.1109/TRO.2008.2006706}
}

@ARTICLE{barr_1981_superquadric,
  author={Barr, Alan H.},
  journal={IEEE Computer Graphics and Applications}, 
  title={Superquadrics and Angle-Preserving Transformations}, 
  year={1981},
  volume={1},
  number={1},
  pages={11-23},
  doi={10.1109/MCG.1981.1673799}
}

@INPROCEEDINGS{park_2024_iwscff_sq,
  author={Park, Tae Ha and Bates, Emily and D'Amico, Simone},
  booktitle={12th International Workshop on Satellite Constellation \& Formation Flying, Kaohsiung, Taiwan}, 
  title={Improving Zero-Shot Abstraction of Unknown Spacecraft 3D Shape as Primitive Assembly}, 
  year={2024},
  volume={},
  number={},
  pages={},
}

@INPROCEEDINGS{bates_2025_gnc_ambiguity,
  author={Bates, Emily and D'Amico, Simone},
  booktitle={48th Rocky Mountain AAS GN\&C Conference}, 
  title={Removing Ambiguities in Concurrent Monocular Single-shot Spacecraft Shape and Pose Estimation Using a Deep Neural Network}, 
  year={2025},
  volume={},
  number={},
  pages={},
}

@inproceedings{caruso_2023_sfm_nerf,
    author={Caruso, Basilio and Mahendrakar, Trupti and Nguyen, Van Minh and White, Ryan T. and Steffen, Todd},
    booktitle={33rd AAS/AIAA Space Flight Mechanics Meeting, Austin, Texas},
    title={{3D} Reconstruction of Non-cooperative Resident Space Objects using {Instant NGP}-accelerated {NeRF} and {D-NeRF}},
    year={2023}
}

@article{muller_2022_acm_instantngp,
   author = {Thomas M\"uller and Alex Evans and Christoph Schied and Alexander Keller},
   city = {New York, NY, USA},
   doi = {10.1145/3528223.3530127},
   issue = {4},
   journal = {ACM Trans. Graph.},
   month = {7},
   publisher = {Association for Computing Machinery},
   title = {Instant Neural Graphics Primitives with a Multiresolution Hash Encoding},
   volume = {41},
   year = {2022},
}

@INPROCEEDINGS{pumarola_2021_cvpr_dnerf,
  author={Pumarola, Albert and Corona, Enric and Pons-Moll, Gerard and Moreno-Noguer, Francesc},
  booktitle={2021 IEEE/CVF Conference on Computer Vision and Pattern Recognition (CVPR)}, 
  title={D-NeRF: Neural Radiance Fields for Dynamic Scenes}, 
  year={2021},
  volume={},
  number={},
  pages={10313-10322},
  doi={10.1109/CVPR46437.2021.01018}
}

@InProceedings{mergy2021_cvprw_nerf,
    author    = {Mergy, Anne and Lecuyer, Gurvan and Derksen, Dawa and Izzo, Dario},
    title     = {Vision-Based Neural Scene Representations for Spacecraft},
    booktitle = {Proceedings of the IEEE/CVF Conference on Computer Vision and Pattern Recognition (CVPR) Workshops},
    month     = {June},
    year      = {2021},
    pages     = {2002-2011},
    doi={10.1109/CVPRW53098.2021.00228}
}

@INPROCEEDINGS{schonberger_2016_cvpr_colmap,
  author={Schönberger, Johannes L. and Frahm, Jan-Michael},
  booktitle={2016 IEEE Conference on Computer Vision and Pattern Recognition (CVPR)}, 
  title={Structure-from-Motion Revisited}, 
  year={2016},
  volume={},
  number={},
  pages={4104-4113},
  doi={10.1109/CVPR.2016.445}}

@InProceedings{schonberger_2016_eccv_colmap,
    author="Sch{\"o}nberger, Johannes L.
    and Zheng, Enliang
    and Frahm, Jan-Michael
    and Pollefeys, Marc",
    title="Pixelwise View Selection for Unstructured Multi-View Stereo",
    booktitle="Computer Vision -- ECCV 2016",
    year="2016",
    pages="501--518",
    doi="10.1007/978-3-319-46487-9_31"
}

@INPROCEEDINGS{barad_2024_isparo_3dgs,
  author={Barad, Kuldeep R and Richard, Antoine and Dentler, Jan and Olivares-Mendez, Miguel and Martinez, Carol},
  booktitle={2024 International Conference on Space Robotics (iSpaRo)}, 
  title={Object-centric Reconstruction and Tracking of Dynamic Unknown Objects Using 3D Gaussian Splatting}, 
  year={2024},
  volume={},
  number={},
  pages={202-209},
  doi={10.1109/iSpaRo60631.2024.10688304}}

@inproceedings{bi_2024_siggraph_gs3,
    author = {Bi, Zoubin and Zeng, Yixin and Zeng, Chong and Pei, Fan and Feng, Xiang and Zhou, Kun and Wu, Hongzhi},
    title = {{GS3}: Efficient Relighting with Triple Gaussian Splatting},
    year = {2024},
    doi = {10.1145/3680528.3687576},
    booktitle = {SIGGRAPH Asia 2024 Conference Papers},
}

@InProceedings{zhang_2024_eccv_gs-w,
author="Zhang, Dongbin
and Wang, Chuming
and Wang, Weitao
and Li, Peihao
and Qin, Minghan
and Wang, Haoqian",
title="Gaussian in the Wild: 3D Gaussian Splatting for Unconstrained Image Collections",
booktitle="Computer Vision -- ECCV 2024",
year="2025",
pages="341--359",
doi="10.1007/978-3-031-73116-7_20"
}

@ARTICLE{wang_2004_tip_ssim,
  author={Zhou Wang and Bovik, A.C. and Sheikh, H.R. and Simoncelli, E.P.},
  journal={IEEE Transactions on Image Processing}, 
  title={Image quality assessment: from error visibility to structural similarity}, 
  year={2004},
  volume={13},
  number={4},
  pages={600-612},
  doi={10.1109/TIP.2003.819861}}

@INPROCEEDINGS{zhang_2018_cvpr_lpips,
  author={Zhang, Richard and Isola, Phillip and Efros, Alexei A. and Shechtman, Eli and Wang, Oliver},
  booktitle={2018 IEEE/CVF Conference on Computer Vision and Pattern Recognition}, 
  title={The Unreasonable Effectiveness of Deep Features as a Perceptual Metric}, 
  year={2018},
  volume={},
  number={},
  pages={586-595},
  doi={10.1109/CVPR.2018.00068}}

@INPROCEEDINGS{tancik_2022_cvpr_blocknerf,
  author={Tancik, Matthew and Casser, Vincent and Yan, Xinchen and Pradhan, Sabeek and Mildenhall, Ben P. and Srinivasan, Pratul and Barron, Jonathan T. and Kretzschmar, Henrik},
  booktitle={2022 IEEE/CVF Conference on Computer Vision and Pattern Recognition (CVPR)}, 
  title={{Block-NeRF}: Scalable Large Scene Neural View Synthesis}, 
  year={2022},
  volume={},
  number={},
  pages={8238-8248},
  doi={10.1109/CVPR52688.2022.00807}}

@INPROCEEDINGS{yang_2023_iccv_crossraynerf,
  author={Yang, Yifan and Zhang, Shuhai and Huang, Zixiong and Zhang, Yubing and Tan, Mingkui},
  booktitle={2023 IEEE/CVF International Conference on Computer Vision (ICCV)}, 
  title={Cross-Ray Neural Radiance Fields for Novel-view Synthesis from Unconstrained Image Collections}, 
  year={2023},
  volume={},
  number={},
  pages={15855-15865},
  doi={10.1109/ICCV51070.2023.01457}}

@INPROCEEDINGS{chen_2022_cvpr_ha-nerf,
  author={Chen, Xingyu and Zhang, Qi and Li, Xiaoyu and Chen, Yue and Feng, Ying and Wang, Xuan and Wang, Jue},
  booktitle={2022 IEEE/CVF Conference on Computer Vision and Pattern Recognition (CVPR)}, 
  title={Hallucinated Neural Radiance Fields in the Wild}, 
  year={2022},
  volume={},
  number={},
  pages={12933-12942},
  doi={10.1109/CVPR52688.2022.01260}}

@InProceedings{dahmani_2024_eccv_swag,
    author="Dahmani, Hiba
    and Bennehar, Moussab
    and Piasco, Nathan
    and Rold{\~a}o, Luis
    and Tsishkou, Dzmitry",
    title="{SWAG}: Splatting in-the-Wild Images with Appearance-Conditioned Gaussians",
    booktitle="Computer Vision -- ECCV 2024",
    year="2025",
    pages="325--340",
    doi="10.1007/978-3-031-73116-7_19",
}

@ARTICLE{engel_2018_tpami_dso,
  author={Engel, Jakob and Koltun, Vladlen and Cremers, Daniel},
  journal={IEEE Transactions on Pattern Analysis and Machine Intelligence}, 
  title={Direct Sparse Odometry}, 
  year={2018},
  volume={40},
  number={3},
  pages={611-625},
  doi={10.1109/TPAMI.2017.2658577}}

@article{montenbruck_2006_ast_eivectorsep,
    title = {{E/I}-vector separation for safe switching of the {GRACE} formation},
    journal = {Aerospace Science and Technology},
    volume = {10},
    number = {7},
    pages = {628-635},
    year = {2006},
    doi = {10.1016/j.ast.2006.04.001},
    author = {Oliver Montenbruck and Michael Kirschner and Simone D'Amico and Srinivas Bettadpur},
}

@misc{kirillov_2023_segmentanything,
      title={Segment Anything}, 
      author={Alexander Kirillov and Eric Mintun and Nikhila Ravi and Hanzi Mao and Chloe Rolland and Laura Gustafson and Tete Xiao and Spencer Whitehead and Alexander C. Berg and Wan-Yen Lo and Piotr Dollár and Ross Girshick},
      year={2023},
      eprint={2304.02643},
      archivePrefix={arXiv},
      primaryClass={cs.CV},
      url={https://arxiv.org/abs/2304.02643}, 
}

@inproceedings{francesch_2025_asc_sq3dgs,
	author={Francesch Huc, Pol and Bates, Emily and D'Amico, Simone}, 
	booktitle={2025 AAS/AIAA Astrodynamics Specialist Conference, Boston, Massachusetts}, 
	title={Fast Learning of Non-Cooperative Spacecraft 3D Models through Primitive Initialization}, 
	year={2025}, 
	volume={}, 
	number={}, 
	pages={}, 
	ISSN={}, 
	month={August 10-14}
}

@INPROCEEDINGS{fu_2024_cvpr_colmapfree3dgs,
  author={Fu, Yang and Wang, Xiaolong and Liu, Sifei and Kulkarni, Amey and Kautz, Jan and Efros, Alexei A.},
  booktitle={2024 IEEE/CVF Conference on Computer Vision and Pattern Recognition (CVPR)}, 
  title={{COLMAP}-Free {3D} Gaussian Splatting}, 
  year={2024},
  volume={},
  number={},
  pages={20796-20805},
  doi={10.1109/CVPR52733.2024.01965}}

@misc{esascifleet,
    title = {{ESA} Science Satellite Fleet},
    howpublished = {Available at \url{https://scifleet.esa.int/}},
    year = {2025},
    note={Accessed 2025/08/01}
}

@misc{starlink_3dmodel,
    title={Starlink Spacex Satellite},
    howpublished = {Available at \url{https://sketchfab.com/3d-models/starlink-spacex-satellite-0a60f6720c5141c9a1c6d71aac108b31}},
    year = {2025},
    note = {Accessed 2025/08/01}
}

\end{document}